\documentclass[journal]{IEEEtran}

\usepackage{multirow}

\ifCLASSINFOpdf
  \usepackage[pdftex]{graphicx}
\else
\fi

\usepackage{amsmath}

\usepackage{algorithmic}

\usepackage{array}
\usepackage[ruled]{algorithm2e}

\ifCLASSOPTIONcompsoc
  \usepackage[nocompress]{cite}
\else
  \usepackage{cite}
\fi
\usepackage{graphicx}
\usepackage{bm}
\usepackage{caption}
\usepackage{hyperref}
\usepackage{amssymb}

\usepackage{ulem}

\usepackage{ragged2e}

\begin{document}
%
\title{Subgraph-aware Few-Shot Inductive Link Prediction via Meta-Learning}
%
%
%

\author{Shuangjia Zheng$^1$,
       Sijie Mai$^1$,
       Ya Sun$^1$,
       Haifeng Hu,
        Yuedong Yang
\IEEEcompsocitemizethanks{
\IEEEcompsocthanksitem Yuedong Yang is with Sun Yat-sen University,
China.\protect\\
E-mail: yangyd25@mail.sysu.edu.cn
\IEEEcompsocthanksitem Haifeng Hu is with Sun Yat-sen University,
China.\protect\\
E-mail: huhaif@mail.sysu.edu.cn
\IEEEcompsocthanksitem Shuangjia Zheng, Sijie Mai, and Ya Sun are with Sun Yat-sen University.\protect\\
E-mail: \{zhengshj9, maisj, suny278\}@mail2.sysu.edu.cn
}
\thanks{ }}

\IEEEtitleabstractindextext{%
\begin{abstract}
\justifying
Link prediction for knowledge graphs aims to predict missing connections between entities. Prevailing methods are limited to a transductive setting and hard to process unseen entities. The recent proposed subgraph-based models provided alternatives to predict links from the subgraph structure surrounding a candidate triplet. However, these methods require abundant known facts of training triplets and perform poorly on relationships that only have a few triplets. In this paper, we propose Meta-iKG, a novel subgraph-based meta-learner for few-shot inductive relation reasoning. Meta-iKG utilizes local subgraphs to transfer subgraph-specific information and learn transferable patterns faster via meta gradients. In this way, we find the model can quickly adapt to few-shot relationships using only a handful of known facts with inductive settings. Moreover, we introduce a large-shot relation update procedure to traditional meta-learning to ensure that our model can generalize well both on few-shot and large-shot relations. We evaluate Meta-iKG on inductive benchmarks sampled from NELL and Freebase, and the  results show that Meta-iKG outperforms the current state-of-the-art methods both in few-shot scenarios and standard inductive settings.
\end{abstract}

}

\maketitle

\IEEEdisplaynontitleabstractindextext

%
\IEEEpeerreviewmaketitle

\section{Introduction}

Knowledge graphs (KGs) are repositories of large amounts of triplets in the form of relations between two entities, encoding knowledge and facts in the world. This kind of graph-structured knowledge has played a critical role across a variety of tasks such as  Semantic Search \cite{xiong2017explicit}, Question Answering \cite{zhang2018link}, and many more. However, due to the limitations of human knowledge and information extraction algorithms, they typically suffer from incompleteness, that is, absent links in the KGs. To automate the KG completion process, numerous latent representation methods have been proposed that condense each entity and relation into a low-dimensional continuous vector space, which can then be utilized to infer missing links by operating the produced embeddings \cite{TransE, tkde_kgembedding2, nickel2015holographic, dismult,trouillon2017knowledge,convE,tkde_kgembedding}.

While these embedding-based models have shown promising performance, prevailing methods typically assume a fixed set of entities in the graph and neglect the evolving character of KGs. However, real-world KGs are often dynamic and ever-evolving \cite{trivedi2017know}, with new entities being added at any given moment, e.g., new users on online shopping platforms. Recently, taking inspiration from the success of graph neural networks (GNNs) in graph structure modeling, a few efforts have been made to subgraph-based inductive relation prediction, combining the beneficial qualities of both scalability and interpretability \cite{grail,mai2020communicative}. The basic strategy behind this type of model is to score a target triplet based on its enclosing subgraph. It can facilitate the prediction of completely novel entities that are not surrounded by known nodes, since the domain-related initial embedding of these emerging entities is not required in the modeling.





Despite the impressiveness of model performance, this framework assumes that there are enough triplets to train robust and effective  reasoning models for each relation in KGs, as the graph neural network usually needs substantial instances to enable the model stability \cite{keriven2020convergence}. However, previous works have shown that a large part of KG relations is actually long-tail \cite{han2018fewrel,xiong2018one} and only occur only a handful of times, which can be referred to as \textit{few-shot} relations. Some pilot experiments have demonstrated that the few-shot scenario incurs the infeasibility of GNN models, resulting in catastrophic performance decline on those few-shot classes \cite{garcia2017few, zhou2019meta}. The inability to handle the presence of very few samples is one of the major challenges for the current GNNs. In fact, few-shot relations often appear when new entities are encountered. Therefore, few-shot inductive relation reasoning is a practical issue of considerable importance that has not been fully resolved.



\textbf{Present Work.} We propose Meta-iKG, a novel subgraph-based meta learner for few-shot inductive relation reasoning. Meta-iKG utilizes local subgraphs to transfer subgraph-specific information and learn transferable patterns faster via meta gradients. Specifically, we first translate link prediction as a subgraph modeling problem. Then, we regard triplet queries with the same relation $r$ in KGs as a single task. Following the previous meta-learning paradigm \cite{finn2017model,li2017meta,tkde_meta} , we use tasks of high-frequency relations to construct a meta-learner, which includes common features across different tasks. The meta-learner can be fast adapted to the tasks of few-shot relations, by providing a good initial point to train their relation-specific subgraph scoring function. Moreover, different from standard meta-learning, we introduce large-shot relation update procedure to eliminate the bias introduced by the few-shot relational meta-updating, which enables our Meta-iKG to generalize well both on large-shot and few-shot relations. We evaluate Meta-iKG on several novel inductive link prediction benchmarks sampled from NELL and Freebase, and experimental results show that Meta-iKG outperforms the state-of-the-art methods both in few-shot and standard inductive settings.


In brief, the main contributions are listed below:
 \begin{itemize}
   \item Introducing a inductive few-shot relation prediction problem which is different from previous work and more suitable for practical scenarios.
   \item Proposing a competitive few-shot inductive KG embedding model, Meta-iKG, that fits the few-shot nature of knowledge graph and can naturally generalize to the unseen entities. Meta-iKG can generalize well both on few-shot and large-shot relations.
   \item Experiments on eight inductive  datasets demonstrate that our model achieves state-of-the-art AUC-PR and Hits@10 across most of them both in few-shot and standard inductive settings.
 \end{itemize}

\section{Related Work}


\uline{\textbf{Inductive relation prediction.}} A close research line is the rule-based approach \cite{galarraga2015fast} and recent proposed differentiable rule learners such as RuleN \cite{meilicke2018fine}, NeuralLP \cite{yang2017differentiable}, and DRUM \cite{DRUM} that simultaneously learn the logical rules and confidence scores in an end-to-end paradigm. Recent studies incorporate graph neural network into inductive relation reasoning to capture multi-hop information around the target triple. GraIL \cite{grail} proposed a subgraph-based relation reasoning framework to process unseen entities and CoMPILE \cite{mai2020communicative} extends the idea by introducing a node-edge communicative message passing mechanism to model the directed subgraphs, which fits the directional nature of knowledge graph and can naturally deal with asymmetric and anti-symmetric relations. Our study can be interpreted as an extension of CoMPILE method to few-shot knowledge graph completion.

\uline {\textbf{Few-shot relation prediction.}} The few-shot learning models mainly fall into two categories: (i) metric based approaches \cite{koch2015siamese,vinyals2016matching} ; (ii) meta-optimizer based approaches \cite{finn2017model,nichol2018first,lee2018gradient} . The former one learns an effective metric and corresponding matching function among a set of training instances.  For example, GMatching  \cite{xiong2017explicit} and MetaR \cite{ chen2019meta} use metric methods to generalize over new relations from a handful of associative relations in a knowledge graph.  They achieves fairly good performance in one-shot relation prediction yet still have two main limitations: (i) they only encode one-hop neighbors of entities and ignore the high-order neighborhood information around a target triplet. (ii) they are limited to transductive settings and cannot process unseen entities. 
Note that MetaR \cite{ chen2019meta} aims to quickly optimize the model parameters given the  gradients on few-shot data instances. One widely-used strategy is the model-agnostic meta-learning (MAML)\cite{finn2017model} which trains model by a small number of gradient updates and leads to fast adapting on a new task. A few attempts have been proposed that combine meta learning with multi-hop reinforcement learning \cite{lv2019adapting} or sequential network \cite{mirtaheri2020one} to perform few-shot link prediction. To the best of our knowledge, this work is the first research on few-shot learning for inductive relation reasoning.

\section{Method}
\subsection{Formulation and Model Overview}

\begin{figure}[h]
  \centering
 \includegraphics[width=0.92\linewidth]{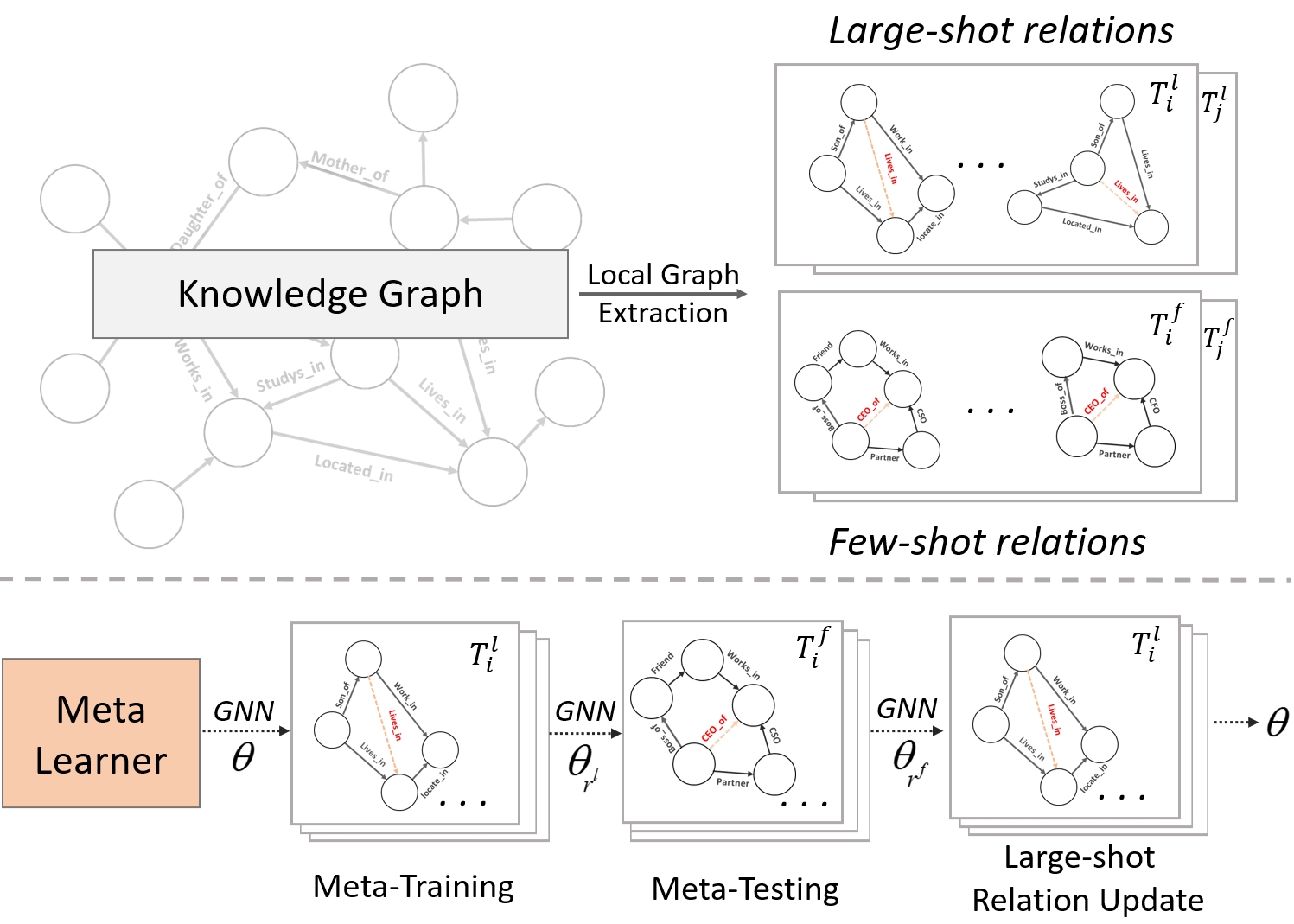}
  \caption{ \label{1111} Overview of Meta-iKG. A) Extracting local enclosing subgraphs around target entities. B) Meta-learner adapts to few-shot and large-shot relations via three-step optimization.}
  \vspace{-0.3cm}
\end{figure}

A triplet in a KG is denoted as $(s, r, t)$ where $s$, $r$, and $t$ refers to the head entity, relation, and tail entity, respectively. Inductive relation prediction aims to evaluate the plausibility of a target triplet $(s_T, r_T, t_T)$ , where the embeddings of $s_T$ and $t_T$ are not available during reasoning. In this paper, to enable the model to generalize well on the relations that only have few training triplets, we split the relations into few-shot and large-shot relations. If the number of triplets including a relation $r$ is fewer than a specific threshold $K_T$, we denote $r$ as a \textit{few-shot relation}, otherwise, it is a \textit{large-shot (normal) relation}. Following the basic idea of meta-learning\cite{finn2017model,xiong2017explicit,chen2019meta,lv2019adapting}, we train triplets with large-shot relations to find well-initialized parameters and adapt the models on triplets with few-shot relations from the found initial parameters. An overview of our Meta-iKG can be seen in Fig.~\ref{1111}. In particular, the proposed learning framework can be divided into two modules: (i) relation-specific learning and (ii) meta-learning. The purpose of relation-specific learning is to learn a GNN model with parameters $\bm{\theta}_r$ for a set of subgraphs surrounding a specific relationship $r$ (task) to identify the transferable patterns. Meta-learning is based on the relation-specific module for learning a meta model with parameters $\bm{\theta}$ and enables fast adaptation for new few-shot tasks. We introduce these two parts in following sections.

\subsection{Subgraph-aware Relation-specific Learning}
For each query relation $r^q$ $\in$ $R$, we learn a relation-specific subgraph scoring function using the triplets with the relation $r^q$ to make predictions. Our subgraph-based scoring function can be divided into three subtasks: (i) extracting the enclosing subgraph surrounding the two target entities, (ii) labeling the entities in the subgraphs with their relative position, and (iii) scoring the labeled subgraphs using GNNs.

\textbf{Subgraph extraction.} We first extract the $h$-hop directed enclosing subgraph $G$ between the target head and tail entities, where $h+1$ is the maximum distance from target head to target tail. This is based on the assumption that the paths connecting the two target entities include the information could infer the target relation \cite{grail}.
Given a triplet $(s, r, t)$, we define $s$ as the 1-hop incoming neighbors of $t$, and $t$ as the 1-hop outgoing neighbor of $s$, and $h$-hop incoming/outgoing neighbors vice verse. Firstly we extract the $h$-hop incoming neighbors for target head  and $h$-hop outgoing neighbors for target tail.
Then we find the common entities of the extracted neighbors of the two target entities. Finally, we build the subgraph by adding the edges (i.e., the triplets) whose heads and tails are in the common entities or target entities.

\textbf{Inductive node labeling function.} Before feeding the $h$-hop subgraph to GNN, we first apply an inductive node labeling function to it, which uses different labels to mark nodes' different roles in the subgraph without leveraging any external domain features and global information. As such, the model has to learn the structural semantics that underlies the extracted subgraph, which is the key reason for the inductive nature of our model. Following \cite{zhang2018link}, we initialize the node embedding $\bm{N}$ by the distances to the target head $s_T$ and tail $t_T$ to embed the relative position of each node in the subgraph.

In detail, for a node $i$ within the subgraph, its label is defined as $\bm{N}_i = \text{one-hot}(d_{si})\oplus \text{one-hot}(d_{it})\in \mathbb{R}^{2(h+1)}$ where $d_{hi}$ denotes the shortest distance from the target head entity to node $i$, and $d_{it}$ denotes the shortest distance from the node $i$ to target tail entity.  Note that similar to GraIL \cite{grail}, the two target nodes, $s_T$ and $t_T$ are uniquely labeled (0,1) and (1,0) so as to be identifiable by the model.

\textbf{Directed Subgraph Scoring:} In principle, our framework can be combined with a wide variety of GNN-based approaches, and here we focus on communicative message passing neural network following the idea of CoMPILE \cite{mai2020communicative}. Formally, given a target triplet $(s_T, r_T, t_T)$ with the enclosing subgraph $G$, the subgraph scoring function can be defined as:
\begin{equation}
\label{eq5}
\setlength{\abovedisplayskip}{3pt}
\setlength{\belowdisplayskip}{3pt}
  \bm{S} = \text{GNN}(G, \bm{N}_{s_T}, \bm{R}_{r_T}, \bm{N}_{t_T}; \bm{\theta} )
\end{equation}
where $\bm{S}$ denotes the predicted plausibility of the target triplet, $\bm{N}_{s_T}$, $\bm{R}_{r_T}$, and $\bm{N}_{t_T}$ denotes the embedding of target head, target relation, and target tail, respectively.
Please refer to  CoMPILE \cite{mai2020communicative} or our Appendix for the detailed procedure. The relation embedding $\bm{R}$ is parameterized as a learnable matrix and is shared across subgraphs.

\subsection{Meta-Learning}
The goal of meta-learning is to learn well-initialized parameters, such that small changes in the parameters will produce significant improvements on the loss function of any task \cite{finn2017model}. In this section, we describe our meta-learning strategy in detail which enables our model generalizes well both on few-shot and large-shot relations. Formally, we consider a meta GNN, i.e. CoMPILE with parameters $\bm{\theta}$. Firstly we divide the relations into few-shot and large-shot relations according to the threshold: $K_T = n_T / n_R \ \times \ \gamma $ where $n_T$ denotes the number of training triplets, $n_R$ denotes the number of relations, and $\gamma$ is a scalar. In each iteration, we sample a batch of relations $r^l$ from large-shot relation set $R^l =\{r|n_r>K_T, r \in R \}$ and a batch of relations $r^f$ from few-shot relation set  $R^f =\{r|n_r\leq K_T, r \in R \}$. Then we sample triplets belonging to $r^l$ and $r^f$ to form the support set $D_S$ and query set $D_Q$, respectively.

The model is firstly updated by the support set $D_S$ in large-shot tasks $T^l$, after which the parameters of the model become $\bm{\theta}_{r^l}$. Following MAML \cite{finn2017model}, the updated parameters $\bm{\theta}_{r^f}$ are computed using one or more gradient descent updates on the few-shot tasks $T^f$. Formally, we will go over many large-shot tasks (i.e., meta-training tasks) in a batch:
\begin{equation}
\label{eq111}
  \bm{\theta}_{r^l} = \bm{\theta} - \bm{\alpha} \cdot \sum_{T^l_i} \nabla_{\bm{\theta}} L_{r^l_i}^{D_S}(\bm{\theta}),
\end{equation}
where $\bm{\alpha}$ is the learning rate, $D_S$ is a support set randomly sampled from the triples belonging to large-shot relations $r^l$ for task $T^l$. After the relation-specific parameters $\bm{\theta}_{r^l}$ is learned, we evaluate $\bm{\theta}_{r^l}$ on the query set $D_Q$ belonging to few-shot relations $r^f$ for task $T^f$. The gradient computed from this evaluation can be used to update the meta policy network CoMPILE with parameters $\bm{\theta}$. Specifically, we update $\bm{\theta}$ using the few-shot tasks $T^f$ (i.e., meta-testing tasks) as follows:
\begin{equation}
\label{eq222}
  \bm{\theta}_{r^f} = \bm{\theta} - \beta \sum_{T^f_i} \nabla_{\bm{\theta}} L_{r^f_i}^{D_Q}(\bm{\theta}_{r^l}),
\end{equation}
where $\beta$ is the meta-learning rate. By this means, Meta-iKG can learn to fast adapt to the few-shot relations with the aim of the well-initialized  parameters $\bm{\theta}_{r^l}$ updated by the support set in large-shot relations.

The above two steps are the regular operations of meta-learning. While promising on the few-shot relations, these traditional operations of meta-learning will introduce bias to the updated parameters, because the final updated parameters $\bm{\theta}_{r^f}$ depend on the few-shot relations and therefore the model may not perform well on large-shot relations. To this end, we innovative to introduce another updating operation on $\bm{\theta}_{r^f}$ using the support set $D_S$  belonging to large-shot tasks $T^l$ with a smaller learning rate $\beta^{'}$, which is called `large-shot relation update procedure'. The equation is shown as below:
\begin{equation}
\label{eq333}
  \bm{\theta} \gets \bm{\theta}_{r^f} - \beta^{'} \sum_{T^l_i} \nabla_{\bm{\theta}_{r^f}}\ L_{r^l_i}^{D_S}(\bm{\theta}_{r^f}),
\end{equation}
where $\bm{\theta}$ is the final updated parameter.
This simple operation enables Meta-iKG to generalize well both on large-shot and few-shot relations. Note that we apply Meta-SGD\cite{li2017meta} as our meta-learner, in which case the learning rate $\bm{\alpha}$ is meta-learnable.
We detail the Meta-iKG in Algorithm 1.


 \begin{algorithm}[t]
 \small
  \caption{Meta-iKG}
  \textbf{Input:} GNN parameter $\bm{\theta}$, learning rate $\beta$ and $\beta^{'}$, meta-learning rate $\bm{\alpha}$, task distribution $p(T^l)$ and $p(T^f)$ \\
  \textbf{Output:} GNN parameter $\bm{\theta}$ \\
 \textbf{step 1}: \
 Initialize $\bm{\theta}$ and $\bm{\alpha}$\\
 \textbf{step 2}: \
 while $not \ done$ do:\\
 \text{\ \ \ \ \ \ \ \ \ \ \ \ \ \ \  } Sampling large-shot tasks $T^l$ from $p(T^l)$\\
 \text{\ \ \ \ \ \ \ \ \ \ \ \ \ \ \  } Obtaining $\bm{\theta}_{r^l}$ using $T^l$, $\bm{\theta}$  and  $\bm{\alpha}$ via Eq.~\ref{eq111}\\
 \text{\ \ \ \ \ \ \ \ \ \ \ \ \ \ \  } Sampling few-shot tasks $T^f$ from $p(T^f)$\\
 \text{\ \ \ \ \ \ \ \ \ \ \ \ \ \ \  } Obtaining $\bm{\theta}_{r^f}$ using $T^f$, $\bm{\theta}_{r^l}$ and  $\beta$ via Eq.~\ref{eq222}\\
  \text{\ \ \ \ \ \ \ \ \ \ \ \ \ \ \  } Updating $\bm{\theta}$ using $T^l$, $\bm{\theta}_{r^f}$ and  $\beta^{'}$  via Eq.~\ref{eq333}\\
 \text{\ \ \ \ \ \ \ \ \ \ \ \ \ \ \ } Updating $\bm{\alpha}$ by:  $\bm{\alpha}\gets \bm{\alpha}$\! -\! $ \beta \sum_{T^f_i} \nabla_{\bm{\alpha}} L_{r^f_i}^{D_Q}(\bm{\alpha})$\\
 \text{\ \ \ \ \ \ \ \ \ \ \ \ \  } end
  \end{algorithm}

\begin{table*}[h]
\centering
\resizebox{1.95\columnwidth}{!}{\begin{tabular}{c|c|c|c|c|c|c|c}
 \hline
   Version  & Train Relations & Train Graph & Train Triplets & Validation Triplets & Test Relations & Test Graph & Test Triplets\\
 \hline
  v1  & 183 & 4,245 & 4,040 & 475 & 146 & 1,993 &  108 \\
 v2  & 213 & 9,739 & 9,462 & 1,142 & 176 & 4,145 & 380 \\
 v3  & 218 & 17,986 & 17,703 & 2,179 & 187 & 7,406 & 779 \\
 v4  & 222 & 27,203 & 26,917 & 1,658 & 204 & 11,714 & 1,369 \\
 \hline
 \end{tabular}}
  \caption{ \label{t44}\textbf{ Inductive FB15k-237 Datasets}}
\end{table*}%

\begin{table*}[h]
\centering
\resizebox{1.95\columnwidth}{!}{\begin{tabular}{c|c|c|c|c|c|c|c}
 \hline
   Version  & Train Relations & Train Graph & Train Triplets & Validation Triplets & Test Relations & Test Graph & Test Triplets\\
 \hline
 v1  & 14 & 4,687 & 3,610 & 379 & 14 & 833 & 81 \\
 v2  & 88 & 8,219 & 7,118  & 921 & 79 & 4,586 & 430 \\
 v3  & 142 & 16,393 &  14,453  & 1,848 & 122 & 8,048 & 686 \\
 v4  & 77 & 7,546 & 6,710 & 419 & 61 & 7,073 & 638 \\
 \hline
 \end{tabular}}
  \caption{ \label{t55}\textbf{ Inductive Nell-995 Datasets}}
\end{table*}%

\begin{table}
\centering
\resizebox{1.0\columnwidth}{!}{\begin{tabular}{c|c|c|c|c|c|c|c|c}
 \hline
  &   \multicolumn{4}{c|}{FB15k-237} &  \multicolumn{4}{c}{NELL-995} \\
 \hline
   Model& \multicolumn{1}{c|}{v1} & \multicolumn{1}{c|}{v2} & \multicolumn{1}{c|}{v3} & v4 & \multicolumn{1}{c|}{v1} & \multicolumn{1}{c|}{v2} & \multicolumn{1}{c|}{v3} & v4\\
 \hline
 RuleN & 79.60 & 82.67 & 83.03 & 84.01 & 67.12 & 80.52 & 73.91 & 77.07\\
 GraIL & 80.45 & 83.66 & 84.35 & 83.08 & 69.35 & 85.04 & \textbf{84.43} & 80.19\\
 CoMPILE & 79.95 & 83.56 & 83.97 & 83.87 & 68.36 & 85.50 & 84.04 & 79.89\\
  \hline
 Meta-iKG (MAML) & 80.31 & 82.95 & 82.52 & \textbf{84.23} & 72.12 & 84.11 & 82.47 & 79.25\\
 Meta-iKG (Meta-SGD) & \textbf{81.10} & \textbf{84.26} & \textbf{84.57} & 83.70 & \textbf{72.50} & \textbf{85.97} & 84.05 & \textbf{81.24}\\
  \hline

 \end{tabular}}
  \caption{ \label{t1}\textbf{ Comparison between Models (AUC-PR).} }
    \vspace{-0.2cm}
\end{table}

\begin{table}
\centering
\resizebox{1.0\columnwidth}{!}{\begin{tabular}{c|c|c|c|c|c|c|c|c}
 \hline
  &   \multicolumn{4}{c|}{FB15k-237} &  \multicolumn{4}{c}{NELL-995} \\
 \hline
   Model& \multicolumn{1}{c|}{v1} & \multicolumn{1}{c|}{v2} & \multicolumn{1}{c|}{v3} & v4 & \multicolumn{1}{c|}{v1} & \multicolumn{1}{c|}{v2} & \multicolumn{1}{c|}{v3} & v4\\
 \hline
  RuleN & 65.35 & 71.68 & 67.84 & 70.53 & 53.70 & 69.77 & 64.29 & 57.92\\
 GraIL & 66.52 & 73.82 & 70.15 & 68.30 & 55.56 & 76.40 & 75.66 & 71.24 \\
 CoMPILE & 66.52 & 72.37 & 69.77 & 70.27 & 62.35 & 76.51 & 75.58 & 68.19 \\
  \hline
 Meta-iKG (MAML) & 66.52 & 72.37 & 68.81 & \textbf{74.32} & 60.49 & 74.07 & \textbf{77.99} & 71.63\\
 Meta-iKG (Meta-SGD) & \textbf{66.96} & \textbf{74.08} & \textbf{71.89} & 72.28 & \textbf{64.20} & \textbf{77.91} & 77.41 & \textbf{73.12} \\
  \hline

 \end{tabular}}
  \caption{ \label{t2}\textbf{ Comparison between Models (Hits@10).} }
    \vspace{-0.2cm}
\end{table}

\begin{table*}[h]
\centering
\resizebox{1.90\columnwidth}{!}{\begin{tabular}{c|c|c|c|c|c|c|c|c|c|c}
 \hline
  &   \multicolumn{3}{c|}{FB15k-237-v1} &  \multicolumn{3}{c|}{FB15k-237-v2} & Nell-995-v1 & \multicolumn{3}{c}{Nell-995-v2} \\
 \hline
  Model& $K\leq5$ & $K\leq10$ & $K\leq K_T$ & $K\leq5$  & $K\leq10$ & $K\leq K_T$ & $K\leq K_T$ & $K\leq5$ & $K\leq10$ & $K\leq K_T$ \\
 \hline
 CoMPILE & 43.75 & 42.86 & 46.43 & 80.08 & 80.00 & 76.92 & 0.00 & 78.18 & 72.41 & 67.14 \\
  \hline
 Meta-iKG (MAML) & \textbf{75.00} & 56.25 & \textbf{57.14} & \textbf{86.67} & 88.33 & \textbf{88.46} & \textbf{50.00} & \textbf{80.00} & 74.36 & 72.50\\
 Meta-iKG (Meta-SGD) & \textbf{75.00} & \textbf{60.71} & 53.57 & \textbf{86.67} & \textbf{90.00} & \textbf{88.46} & \textbf{50.00} &  78.00 & \textbf{78.21} & \textbf{76.25} \\
  \hline
 \end{tabular}}
  \caption{ \label{t5}\textbf{ Comparison on Few-shot Relations (Hits@10).} For Nell-995-v1, since there is no relation whose number of training triplets is fewer than 10, we only present the result of $K\leq K_T$ setting.}
  \vspace{-0.3cm}
\end{table*}%

\section{Experiments}
\subsection{Dataset}
FB15K-237 \cite{toutanova2015representing} and NELL-995 \cite{xiong2017explicit} are two commonly used databases for link prediction. We follow CoMPILE \cite{mai2020communicative} to extract the inductive datasets which have filtered out the triplets that have no enclosing subgraph between the target entities under hop $h$ and ensure more accurate evaluation of the models.  The statistics of the filtered inductive FB15k-237 and Nell-995 datasets are shown in Table~\ref{t44} and Table~\ref{t55}, respectively. Note that CoMPILE extracts three versions of inductive datasets for each dataset, while we extract a new version of inductive datasets for FB15k-237 and Nell-995 (i.e., the v4), respectively. The inductive datasets are the filtered versions of the corresponding inductive datasets extracted in GraIL \cite{grail}. Since we use more challenge inductive datasets, so the performance of the models is generally weaker than the performance reported on GraIL \cite{grail}.


The subgraphs of train triplets and validation triplets are extracted from the train graph, and the subgraphs of the  test triplets are extracted from the test graph. The test graph contains entities that do not presented in the train graph, and the train graph contains all the relations in the test graph. The size of the train graph refers to the number of the triplets contained in the train graph, and the size of test graph vice verse. 
Note that the train triplets are all in the train graph, but the test triplets are not in the test graph. 
Also, the validation triplets are not in the train graph.
These settings are the same as those in GraIL and CoMPILE.


To perform the Hits@10 experiment, we need to sample 49 negative head triplets and 49 negative tail triplets for each test triplet (this setting is consistent with GraIL and CoMPILE; the GraIL paper said they sampled 50 negative triplets, but actually it is 49 in the code), where the negative head/tail triplets are generated by replacing the head/tail of the test triplet with other entities. Therefore, we do not use the test triplets that cannot find enough negative head or tail  triplets to evaluate the model. We evaluate our model both on original inductive setting and on few-shot inductive setting where the relations whose number of training triplets (denoted as $K$) is fewer than 5, 10, or $K_T$ are selected for testing.

\subsection{Experimental Details}
\textbf{Evaluation Protocol}: To be consistent with the prior methods, we use AUC-PR and Hits@10 to evaluate the models. To compute AUC-PR, for each test triplet, we sample one negative triplet and evaluate which triplet has larger score. For Hits@10, we rank each test triplet among the sampled 49 negative head/tail triplets and evaluate whether test triplet score makes it into top 10. The negative triplets are obtained by replacing the head or tail of the test triplets with other entities. 
We train the model for four times and average the testing results to obtain the final performance.

\textbf{Hyper-parameter Setting}: We implement our model on Pytorch. We run the model for 20 epochs in the original inductive datasets, and each epoch contains 100 meta-updates.  We use Adam \cite{Kingma2014Adam} as optimizer with learning rate being 0.001. The hop number $h$ is set to 3 which is consistent with GraIL and CoMPILE.  The number of iterations $l$ is set to 3 or 4 which depends on the datasets. The few-shot factor $\gamma$ is set to 0.1.  For the sake of complexity, the subgraph will be pruned if it contains too many nodes, and we ensure that the pruned subgraph can also have a complete path between the target head and target tail. The codes and datasets will be publicly available once accepted.

\subsection{Baselines}
We compare  Meta-iKG with the state-of-the-art subgraph-based inductive models GraIL \cite{grail} and CoMIPLE\cite{mai2020communicative} as well as the rule-based algorithm RuleN \cite{meilicke2018fine}. For meta-learning strategy, we use both MAML \cite{finn2017model} and Meta-SGD \cite{li2017meta} as two different meta-learning strategies to create model variants.

\label{sec:one}

\subsection{Comparisons on Inductive Datasets.} We compare our proposed Meta-iKG with other baselines on the inductive datasets.  From the Table~\ref{t1} and Table~\ref{t2}, we can conclude that:
(a) our Meta-iKG, especially the Meta-SGD version, achieves the best performance on the majority of the inductive datasets in terms of both the AUC-PR and Hits@10 evaluation metrics by a significant margin;
(b) the MAML version of our Meta-iKG performs worse than the Meta-SGD version, demonstrating the importance of the meta-trainable learning rate in the fast adaption of meta-learning.
These results demonstrate the effectiveness of our unique meta-learning strategy in inductive relation prediction, which can generalize well both on large-shot and few-shot relations via the updating procedure in Eq.~\ref{eq111}, Eq.~\ref{eq222} and Eq.~\ref{eq333}. Our model enables a better prediction capacity on few-shot relations without sacrificing the performance on the overall datasets.



\subsection{Comparison on Few-Shot Testing Sets.} In this section, we evaluate our Meta-iKG on the few-shot relations, where the relations whose number of training triplets $K$ is fewer than 5, 10, and $K_T$ are selected to evaluate the stability and robustness on different settings of few-shot inductive relation prediction ($K_T$ is the threshold during training to split the large-shot and few-shot relations). For comparison, we also present the results of CoMPILE, which is the baseline message passing model for our Meta-iKG. According to Table~\ref{t5} and ~\ref{t555}, we find that (a) both versions of Meta-iKG significantly outperform the baseline CoMPILE on the majority of settings, and the improvement is much larger compared to that in the original inductive datasets; (b) although the MAML version of Meta-iKG performs weaker than CoMPILE on some original datasets, it manages to outperform CoMPILE significantly on the few-shot scenarios.
These results demonstrate the effectiveness and stability of our meta-learning strategies in few-shot inductive relation prediction by learning transferable patterns of subgraphs faster via meta gradients.

\begin{table*}[!htb]
\centering
\resizebox{1.90\columnwidth}{!}{\begin{tabular}{c|c|c|c|c|c|c|c|c|c|c}
 \hline
  &   \multicolumn{3}{c|}{FB15k-237-v3} &  \multicolumn{1}{c|}{FB15k-237-v4} & \multicolumn{3}{c|}{Nell-995-v3} & \multicolumn{3}{c}{Nell-995-v4} \\
 \hline
 Model & $K\leq5$ & $K\leq10$ & $K\leq K_T$  & $K\leq K_T$ & $K\leq5$  & $K\leq10$ & $K\leq K_T$ & $K\leq5$ & $K\leq10$ & $K\leq K_T$ \\
 \hline
 CoMPILE & 58.33 & 57.14 & 70.00  &  68.57 & \textbf{65.87}  & 70.95 & 70.25  & 72.50 & 72.20 & 72.55 \\
  \hline
 Meta-iKG (MAML) & \textbf{66.67} & \textbf{64.29}  & \textbf{80.00} & \textbf{76.42} & \textbf{65.87} &  71.43 & 71.07 & 74.25 & 73.90 & 74.34 \\
 Meta-iKG (Meta-SGD) & 58.33 & 57.14 & 73.33  & 73.57 & \textbf{65.87} & \textbf{72.86} & \textbf{73.14} &  \textbf{78.63} & \textbf{78.29} & \textbf{78.28} \\
 \hline
 \end{tabular}}
  \caption{ \label{t555}\textbf{ Comparison on Few-shot Relations (Hits@10).} For FB15k-237-v4, since there are few testing triplets that belong to the relations whose number of training triplets is fewer than 10, we only present the $K\leq K_T$ case.}
  \vspace{-0.2cm}
\end{table*}%

\subsection{Analysis on the Model Complexity}
Since the meta-learning algorithm does not introduce additional module to process the enclosing subgraph but merely modifies the optimization algorithm of the model, the time complexity remains unchanged compared to the original model during prediction. However, due to the flexible size of the subgraph, the common proxy to evaluate the time complexity of the model, i.e., FLOPs (which computes the number of operations go through in the forward pass given a sample), is changeable and depends on the size of the subgraph. As for the space complexity, for the MAML version of our Meta-iKG, since it introduces no additional parameters, the number of parameters is the same as that of the original CoMPILE (which is 41,185 when the dimensionality of relation embedding $d$ is set to 32 and the number of layers $l$ is set to 3). For the Meta-SGD version of our Meta-iKG, since it introduces learnable learning rate parameter $\bm{\alpha}$ which has the same size as the parameter of the GNN (i.e., CoMPILE), it has twice as many parameters as that of CoMPILE. Notably, the learning rate parameter $\bm{\alpha}$ is not added to the model itself, but is used to optimize the model.

\subsection{Analysis on the Influence of $K$}
\begin{figure}[h]
  \centering
 \includegraphics[width=0.8\linewidth]{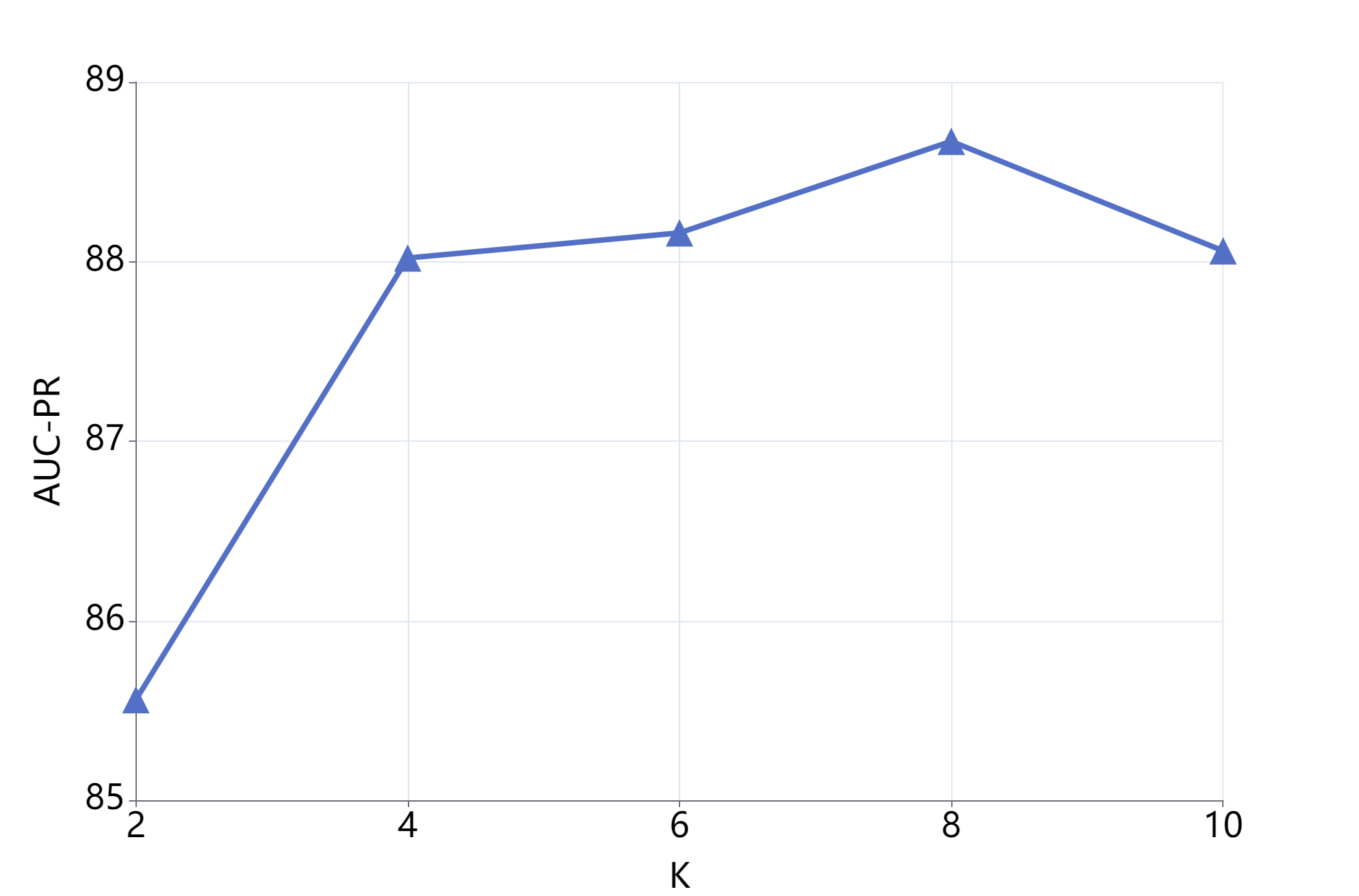}
  \caption{\label{fig22}Analysis on the Influence of $K$.}
  \vspace{-0.3cm}
\end{figure}

We study the effect of the number of training triplets $K$ per relation on the performance of FB15k-237-v2 dataset. We select the relations whose number of training triplets is no less than and no larger than $K_T$ ($10\leq K \leq K_T$), and some training triplets are randomly removed until there are only $K$ triples per selected relation. $K$ is selected from 2 to 10 for each relation in our experiments, and we also evaluate the predictive result of these relations without removing any triplets for comparison. From the results in Fig.~\ref{fig22} we can infer that when $K\geq 6$, the performance of the model tends to converge (the performance has no significant difference with the non-removed version whose AUC-PR is 88.89), suggesting that Meta-iKG can reach satisfactory performance with a small number of training samples. These results further demonstrate the effectiveness of our meta-learning method in few-shot inductive relation prediction.

\subsection{Ablation Studies on Meta-learning}
We investigate the effectiveness of the introduced large-shot relation update procedure and the relation split operation. As presented in the `W/O LRUP' case on Table~\ref{t333}, when the large-shot relation update procedure is removed, significant drop on performance is observed, which suggests that the model cannot generalize well on the standard inductive datasets. Note that the model still performs well on few-shot relation (the Hits@10 is 62.5\% on FB15k-237-v1 dataset when $K\leq K_T$), which support our claim that the model cannot perform well on large-shot relation using the regular meta-learning procedure.  In the case of  `W/O PRO', we do not split the relations into large-shot and few-shot relations, and randomly sample a batch of relations for meta-training and meta-testing during each iteration. Without PRO, the results show that there is a decline of performance on both the datasets (over 2.5\% drops on AUC-PR of FB15k-237-v1 dataset), which demonstrates the effectiveness of our strategy that adapting the model trained on large-shot relations to few-shot relations.

\begin{table}[!htb]
\centering
\resizebox{.98\columnwidth}{!}{\begin{tabular}{c|c|c|c|c}
 \hline
    & \multicolumn{2}{c|}{FB15k-237-v1} & \multicolumn{2}{c}{FB15k-237-v2}\\
 \hline
  & AUC-PR & Hits@10  & AUC-PR & Hits@10\\
 \hline
W/O LRUP & 63.97 & 43.48 & 63.37  & 39.08 \\

W/O RPO & 78.34 & 65.22 &  84.14& 73.42\\

 \hline
 Meta-iKG & \textbf{81.10} & \textbf{66.96} & \textbf{84.26} & \textbf{74.08}\\
 \hline
 \end{tabular}}
  \caption{ \label{t333}\textbf{ Ablation Studies on Inductive Datasets.} The `LRUP' refers to our designed large-shot relation update procedure, and `RPO' represents relation split operation.}
  \vspace{-0.3cm}
\end{table}%

\section{Conclusion}
We present Meta-iKG, a novel method for few-shot inductive relational inference. Meta-iKG uses local subgraphs to convey subgraph-specific information and to learn transferable patterns faster via meta-gradients.  We evaluate Meta-iKG on two novel several-shot inductive link prediction benchmarks, and the experimental results show that Meta-iKG outperforms state-of-the-art methods.

\ifCLASSOPTIONcaptionsoff
  \newpage
\fi

\bibliographystyle{IEEEtran}


%

%

\bibliography{sentiment2}




\end{document}